\documentclass[letterpaper, 10 pt, conference]{ieeeconf}
\IEEEoverridecommandlockouts
\usepackage{cite}
\usepackage{amsmath,amssymb,amsfonts}
\usepackage{caption} 

\usepackage[normalem]{ulem}
\usepackage{amssymb}
\usepackage{pifont}
\newcommand{\cmark}{\ding{51}}%
\usepackage[table,xcdraw]{xcolor}
\usepackage{algorithmic}
\usepackage{graphicx}
\usepackage{soul}
\usepackage{tabularx}
\usepackage{textcomp}
\usepackage{xcolor}
\usepackage{booktabs}
\usepackage[hyphens]{url}
\usepackage[colorlinks]{hyperref}
\def\BibTeX{{\rm B\kern-.05em{\sc i\kern-.025em b}\kern-.08em
    T\kern-.1667em\lower.7ex\hbox{E}\kern-.125emX}}
\begin{document}

\title{A Modular Framework for Flexible Planning in Human-Robot Collaboration\\

\thanks{$*$ Corresponding author \tt\small {valerio.belcamino@edu.unige.it}}
\thanks{$^{1}$ TheEngineRoom, Department of Informatics Bioengineering, Robotics and System Engineering, University of Genoa, Genoa, Italy.}
\thanks{$^{2}$ Cognitive Architecture for Collaborative Technologies Unit (CONTACT), Italian Institute of Technology, Genoa, Italy }%

}
\DeclareRobustCommand{\IEEEauthorrefmark}[1]{\smash{\textsuperscript{\footnotesize #1}}}

\author{
Valerio Belcamino$^{*,1}$,
Mariya Kilina$^{1}$, Linda Lastrico$^{2}$, Alessandro Carfì$^{1}$, Fulvio Mastrogiovanni$^{1}$\thanks{© 2024 IEEE.  Personal use of this material is permitted.  Permission from IEEE must be obtained for all other uses, in any current or future media, including reprinting/republishing this material for advertising or promotional purposes, creating new collective works, for resale or redistribution to servers or lists, or reuse of any copyrighted component of this work in other works.}}


\maketitle

\begin{abstract}
This paper presents a comprehensive framework to enhance Human-Robot Collaboration (HRC) in real-world scenarios. It introduces a formalism to model articulated tasks, requiring cooperation between two agents, through a smaller set of primitives. Our implementation leverages Hierarchical Task Networks (HTN) planning and a modular multisensory perception pipeline, which includes vision, human activity recognition, and tactile sensing.
To showcase the system's scalability, we present an experimental scenario where two humans alternate in collaborating with a Baxter robot to assemble four pieces of furniture with variable components. 
This integration highlights promising advancements in HRC, suggesting a scalable approach for complex, cooperative tasks across diverse applications.
\end{abstract}

\section{Introduction}\label{sec:intro}
With the advancement of technology, Human-Robot Interaction (HRI) has emerged as a central research area at the intersection of robotics, artificial intelligence, and cognitive science. The interaction between humans and robots often involves collaboration, aiming to combine the best humans and robots skills to enhance productivity, efficiency, and versatility across numerous applications. These applications range from manufacturing\cite{10.1007/978-3-030-46212-3_2} and logistics\cite{logistics} to healthcare\cite{healthcare} and domestic assistance\cite{qin2023multimodal}. For the collaboration to be successful, the robot should be able to perceive the environment, and plan and act to accommodate human needs and objectives. 

\begin{figure}[!ht]
\centering
\includegraphics[width=\columnwidth]{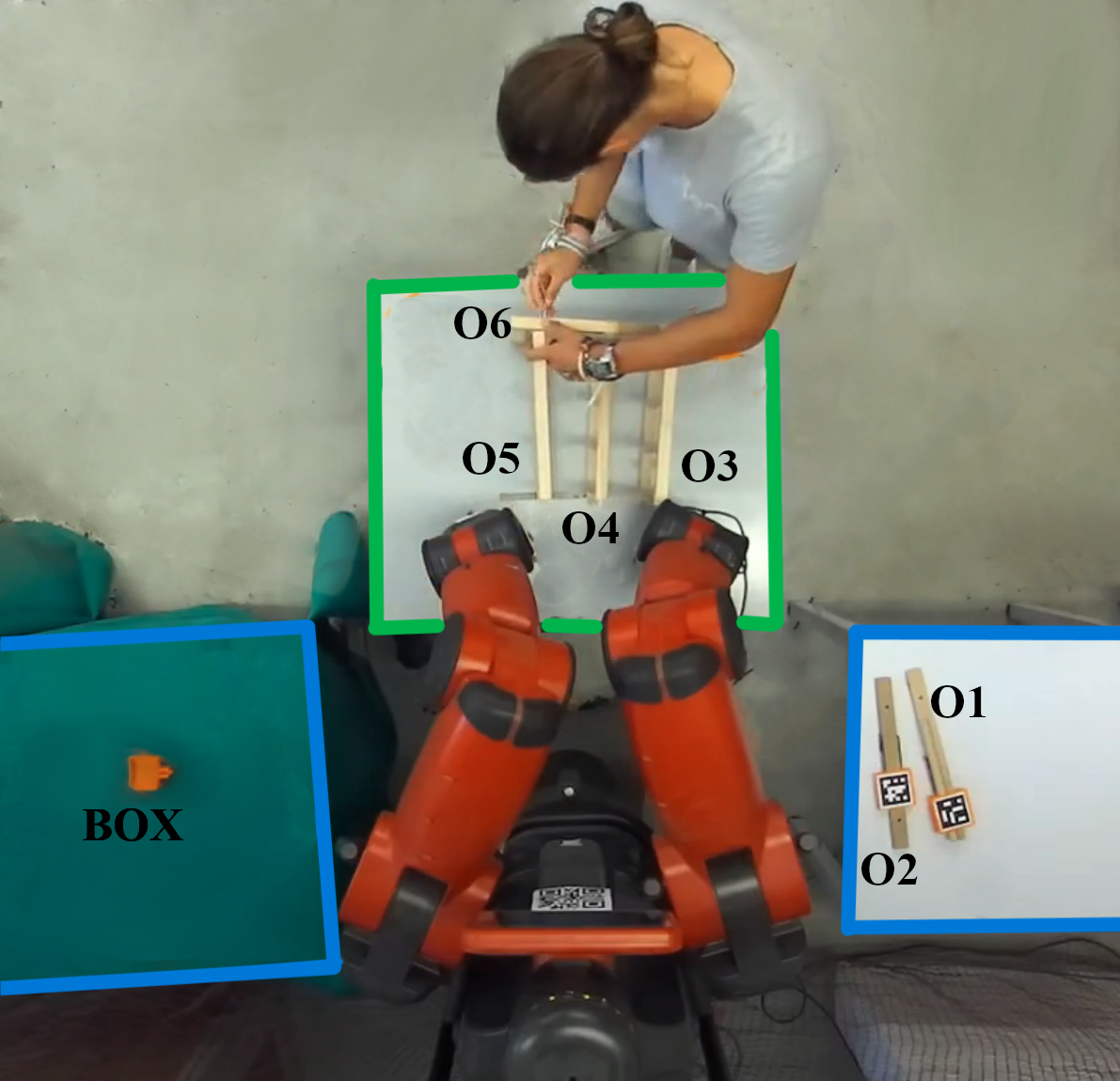}
\caption{A top view of the experimental scenario, in which the collaborative robot Baxter waits for the human to assemble the furniture pieces before continuing the interaction. We show in blue the robot workspace and in green the shared workspace. The labels O1 to O6 point to the components needed for the assembly.}
\label{fig:setup}
\end{figure}

Coordination is the cornerstone of effective collaboration in a multi-agent scenario \cite{coordination1, coordination2}, as it allows for minimizing each participant's idle time and decreases the global duration of the tasks. Coordination is influenced by several factors, such as the agents' dexterity and their sensorial capabilities. However, we focus our analysis on the perception, since the dexterity is strictly coupled to the kinematic structure of the agent and is out of the scope of our work.
In the case of Human-Robot Collaboration, to better understand how the coordination is influenced by the sensorial abilities of the two agents, it is necessary to describe their abilities separately. More specifically, we can distinguish three important aspects: self-perception, environment perception, and the perception of the coworker.

Humans have excellent capabilities in terms of both movement and perception. For example, they can estimate dynamically their position in space through a multi-modal ability called proprioception. Moreover, they mainly rely on vision and touch to gather information about the environment and nearby objects, enabling them to navigate and interact smoothly with their surroundings. Lastly, there is the ability to read the intentions of others, which, in human-human collaboration, is achieved through various communication means, including gestures, posture, gaze, and speech. HRC architectures aim to achieve the same ability resorting to gestures \cite{TELLAECHE201857}, mixed reality \cite{maccio, 10.1007/978-3-030-99188-3_21}, digital twins \cite{simoneMohamad}, voice interfaces \cite{10.1145/3203305} and non-verbal cues \cite{mutlu2009nonverbal, calisgan2012identifying, miwa}, or by making robot's motions more communicative \cite{lastrico_if_2022}.

On the robot side, posture is determined by the rotation of the joints, which is precisely measurable, while the perception of the environment and the coworkers relies on multiple types of sensory modules that can be integrated. Furthermore, the robot should be able to anticipate, or at least recognize, the upcoming actions of the human partner. This predictive ability can facilitate action planning to avoid collisions and optimize the robot’s effectiveness. Drawing upon workspace observations and predictions of future human actions, the robot must dynamically select the most appropriate action online. This real-time adaptation is crucial since pre-planned actions may not align with the evolving situation. Finally, the robot needs to execute actions in ways that align with human needs and conform to safety constraints, thereby ensuring a secure interaction.

We examined recent literature on HRC frameworks aiming to uncover common features contributing to the success and effectiveness of these solutions.  Throughout this analysis, we identified five core features that contribute to the effectiveness of a framework:
\begin{itemize}
\item[]\begin{itemize}
    \item [P1] Modular Architecture
    \item [P2] Runtime Flexibility
    \item [P3] Extensible Task Description
    \item [P4] Low Computational Overhead
    \item [P5] Interpretability
\end{itemize}
\end{itemize}

P1 refers to the \textit{modularity} of the system. More in detail, the architecture of the HRC framework must ensure the ability to model collaboration among a variable number of agents, either human or robot \cite{app122412645, johannsmeier2016hierarchical}. Modularity is crucial, as it allows to model dynamically collaborative scenarios varying over time. Additionally, the framework must be designed to easily accommodate the integration, removal or substitution of modules within the perception pipeline \cite{ren2021digital, miwa}, ensuring adaptability and customization. This allows for the extension of frameworks to new domains or the addition of new functionalities, unlike ML-based frameworks which often require extensive retraining \cite{zhang2022reinforcement, cheng2020towards}.

P2 emphasizes \textit{runtime flexibility} within the HRC framework and requires the ability to adapt the planned action to accommodate the different conditions that may occur during the collaboration. In particular, the system should be flexible enough to adapt to different human behaviours and recover from potential errors. To support this, systems often integrate knowledge about agents' expertise and allocate tasks accordingly \cite{pupa2022resilient}, ensuring efficient task execution and error handling. However, not all frameworks support online re-planning \cite{evangelou2021approach, lee2022task} due to computational constraints, particularly those utilizing techniques with high computational overhead such as digital twin technology \cite{ren2021digital}. Despite its potential benefits, offline planning followed by execution is often sub-optimal due to its inability to respond dynamically to errors or changing conditions.

P3 emphasizes the framework's capacity for expansion and adjustment via an \textit{Extensible Task Description}. This task description should feature thoughtfully chosen general action templates that can be extended to accommodate varying hardware architectures. This characteristic enhances the framework's versatility by enabling updates to the task description and the modelling of problems from diverse domains. Moreover, this aspect aligns with the findings in the literature, where hierarchically structured task planners are favoured \cite{ren2021digital, pupa2022resilient, cesta2016towards, johannsmeier2016hierarchical} for their ability to break down tasks into smaller, more manageable steps — a strategy that resonates with human thinking patterns and facilitates adaptation to different scenarios. Additionally, P3 complements P1, as the careful selection of templates promotes the modularity of the system.

P4 underlines the importance of a low \textit{computational overhead}. The framework should have a low impact on system resources at runtime, ensuring effective performance without weighing on the application. Solutions that solely plan actions offline, lacking support for online re-planning or error handling, naturally incur minimal computational overhead. However, the absence of these vital functionalities compromises the system's effectiveness. The main metrics used to assess the system's computational load are the overall task execution time \cite{app122412645, zhang2022reinforcement}, as well as the execution time for each action \cite{han2021cognitive} and the idle time of agents \cite{han2021cognitive, zhang2022reinforcement}.

Lastly, P5 emphasizes the importance of providing transparent and coherent \textit{justification} for the framework's actions, promoting user comprehension and trust. In this aspect, machine learning-based approaches, such as reinforcement learning \cite{zhang2022reinforcement} and LSTM \cite{cheng2020towards}, often exhibit sub-optimal performance due to their lack of explainability. Conversely, hierarchically structured solutions like HTNs \cite{buisan2021human, tewari2021towards, favier2022robust, stramandinoli2019affordance} and AND/OR graphs \cite{pupa2022resilient, johannsmeier2016hierarchical} offer clear and easily understandable reasoning. Other methods employ node-based graphical interfaces \cite{petzoldt2022implementation} to visualize the sequence of actions in their planners for enhanced clarity. Another viable option is the Hidden Markov Model \cite{zhang2024early}, which aids in tracking decision-making processes based on environmental observations.

Drawing from observations in the literature, it becomes evident that hierarchically structured planners, particularly HTNs, are predominantly favoured for human-robot collaboration frameworks. This preference stems from their ability to address the properties outlined above comprehensively. While certain implementations may deviate and lack one or more of these properties — such as opting for separate planners for each agent \cite{favier2022robust}, which sacrifices modularity and increases computational overhead — HTNs remain overall a compelling choice. As such, we have selected them for our framework and endeavour to leverage their strengths to fulfil all five features efficiently.

This work aims to present an extensible framework for human-robot collaboration considering the properties we identified.
Additionally, we present an implementation of our framework designed for a human-robot collaborative assembly scenario. We describe the choice of the action templates, their implementations in a specific hardware architecture, the planning pipeline with the corresponding state variables and the perception pipeline.
Finally, to showcase the adaptability of our framework to diverse tasks, we designed an experimental setup involving the assembly of four distinct pieces of furniture. This scenario serves as a proof of concept, demonstrating our framework's adherence to the five properties previously outlined. Additionally, we provide quantitative metrics concerning fluency \cite{Hoffman}, including robot idle time, human idle time, functional delay, and concurrent action time.

\begin{table*}[]
\centering
\resizebox{\textwidth}{!}{%
\bgroup
\def\arraystretch{1.5}
\begin{tabular}{cccccccccccccc}

\multicolumn{1}{l}{} & \multicolumn{6}{c}{\textbf{Preconditions}} & \multicolumn{1}{l}{} & \multicolumn{6}{c}{\textbf{Effects}} \\ \cline{2-7} \cline{9-14} 
\multicolumn{1}{l}{} & \multicolumn{1}{c}{Grasp} & \multicolumn{1}{c}{Release} & \multicolumn{1}{c}{Move} & \multicolumn{1}{c}{Manipulate} & \multicolumn{1}{c}{Wait} & \multicolumn{1}{c}{Perceive} & \multicolumn{1}{c}{} & \multicolumn{1}{c}{Grasp} & \multicolumn{1}{c}{Release} & \multicolumn{1}{c}{Move} & \multicolumn{1}{c}{Manipulate} & \multicolumn{1}{c}{Wait} & Perceive \\ \cline{1-7} \cline{9-14} 
\multicolumn{1}{l}{\begin{tabular}[c]{@{}c@{}}End-Effector \\ Availability (EEA)\end{tabular}} & \cmark & \cmark &  & \cmark &  & \multicolumn{1}{c}{} & \multicolumn{1}{c}{} & \cmark & \cmark &  &  &  &  \\ \cline{1-7} \cline{9-14} 
\multicolumn{1}{l}{Agent Pose (AP)} & \cmark &  &  &  &  & \multicolumn{1}{c}{} & \multicolumn{1}{c}{} & \cmark & \cmark & \cmark & \cmark &  &  \\ \cline{1-7} \cline{9-14} 
\multicolumn{1}{l}{Object Pose (OP)} & \cmark &  &  & \cmark &  & \multicolumn{1}{c}{} & \multicolumn{1}{c}{} &  &  & ? & \cmark &  &  \\ \cline{1-7} \cline{9-14} 
\multicolumn{1}{l}{Object Characteristics (OC)} & \cmark &  &  & \cmark &  & \multicolumn{1}{c}{} & \multicolumn{1}{c}{} & \cmark & \cmark &  & \cmark &  &  \\
\bottomrule
\end{tabular}%
\egroup
}
\caption{Description of the interplay of actions and state features. On the left are the preconditions to be met to execute the actions; on the right are the state features (effects) that get updated after the given action. The symbol \textit{`V'} in the Preconditions section means that a certain action requires the corresponding state variable to respect some constraints. Instead, for the Effects, it means that the action updates the state variable. The symbol \textit{`?'} describes an action that can update a feature only if some additional requirements are met. In this case, Move can update the Object Pose feature only if End-Effector Availability was False when the action began. }
\label{tab:state-actions}
\end{table*}

\section{Problem Formalization}
\label{Formalism}
For a robot to collaborate effectively with a human, it must understand the state of the interaction and act accordingly. This problem can be formalized by introducing two concepts: the interaction state and the task plan. The interaction state includes all relevant variables necessary to describe the collaborative task and its effect on the environment, for example, the positions of objects and the poses of involved agents. On the other hand, the task plan details the actions that must take place to reach a desired goal, which is a specific interaction state. Regardless of who performs it, each action affects the state, and their cumulative effect should lead to the final goal state. However, as the state has a preconditioning effect on the actions, only those whose preconditions are met in the state can be performed; for instance, an agent must grasp a tool before using it. Therefore, an accurate task planning solution for human-robot collaboration should be able to choose between feasible actions, evaluating their effect on the environment and considering that the human counterpart could, in the meantime, alter the state.

In human-robot collaboration, the state description, the set of actions in the plan, and their allocation to agents are influenced by various factors, such as the goal, the number of agents involved, and the limitations of the robotic platform \cite{robotics8040100}. For these reasons, providing an extensive and general formalization of the problem is difficult. In this section, we tackle the problem by describing general state properties and actions and how their interaction can be used to carry on the collaboration. This formalization has been empirically synthesized and will be tested in a practical scenario in the following sections. It is worth noting that although this formalization is provided in the HRC context, it can also be used to describe planning problems where a single agent is involved.

In our formalization, we have identified four state features that describe the interaction of agents with the environment. These features are end-effectors availability (EEA), agents' pose (AP), objects' pose (OP), and objects' characteristics (OC), where each feature is represented as a vector containing a feature descriptor for each entity of interest. 
The agents' pose (AP) contains the agents' joint states and positions in space. Given the unique joint configuration of an agent, this feature allows for determining the pose in the space of each of their links, including the end-effector. The objects' pose (OP) contains the objects' positions, orientations in space, and the occasional internal degree of freedom that some objects may have. Finally, the objects' characteristic (OC) is a generic container used to collect the unique characteristics of the objects. For example, a container could have a characteristic describing if it is full or empty, while a tool might have one concerning the most appropriate grasping type. Hence, the OC allows for describing multiple and dissimilar characteristics of each object.
For the assembly task shown in Fig. \ref{fig:setup}, the state will contain:

\begin{equation*}
\begin{gathered}
EEA = \{eea_{H,r}, eea_{H,l}, eea_{R,r}, eea_{R,l}\} \\
AP = \{ap_{H}, ap_{R}\} \\
OP = \{op_{box}, op_{O1}, op_{O2}, op_{O3}, op_{O4}, op_{O5}, op_{O6}\} \\
OC = \{oc_{box}, oc_{O1}, oc_{O2}, oc_{O3}, oc_{O4}, oc_{O5}, oc_{O6}\}
\end{gathered}
\end{equation*}

where $eea_{H,r}$ and $eea_{H,l}$ describe the availability of the human hands and $eea_{R,r}$, $eea_{R,l}$ describe the availability of the robot grippers. Additionally, $ap_{H}$, $ap_{R}$ contain the joint states of the two agents as well as their location and orientation in the world frame. As for OP, it holds an entry for each of the five wooden pieces and one for the box of screws; each entry contains the location and orientation of the corresponding object. Similarly, OC considers a characteristic for each object; $oc_{box}$ shows that the box has been emptied, while $oc_{O1}$ and $oc_{O2}$ represent the two pieces located in the robot workspace, each characterized by the attribute of being "not grasped". Finally, $oc_{O3} - oc_{O6}$ refer to the pieces located in the shared workspace and their characteristics show that they have been assembled.

Having described the state variables we can introduce our choice for the actions; we have identified six actions that comply with the requirements set by P3 being generic enough to describe different collaborative assembly scenarios and allowing the development of specific implementations based on the hardware architecture.
These templates are: \textit{grasp}, \textit{release}, \textit{move}, \textit{manipulate}, \textit{wait}, and \textit{perceive}.
\textit{Grasp} and \textit{release} involve opening and closing the end-effector, respectively. \textit{Move} refers to any agent motion unrelated to the end-effector. \textit{Manipulate} describes fine motions performed by the end-effector to alter the state of an object, such as activating a power tool. \textit{Wait} and \textit{perceive} are two actions that do not directly affect the environment. \textit{Wait} refers to instances where the agent must wait for an action precondition to be satisfied by a change in the state. \textit{Perceive} allows the agent to update its internal representation of the environment state. Of course, not every agent can perform all actions. For example, a robot with a gripper-like end-effector cannot manipulate objects, and a fully passive robot can only perceive the scene. 

As mentioned earlier, each action affects the collaboration state and has preconditions that must be met before its execution. Table \ref{tab:state-actions} highlights the state features that each action template affects and the state feature that could impose a precondition for execution. For example, the \textit{Release} action requires to satisfy a precondition on the EEA (i.e., the agent is supposed to be holding something) and affects the EEA, the AP, and the OC. The first two features are updated considering the current state of the gripper, while, in this case, OC refer to the object property of not being grasped by any agent.

Considering the proposed interaction between state and actions, we can model multiple collaborative tasks. Indeed, given the appropriate description of the state evolution and associated actions, a single agent or a team can execute the resulting plan. In the latter case, actions should be assigned to the team members based on their capabilities. As such, it is necessary to encode the proposed formalism within a framework that supports online implementation and allows for the differentiation of agents' skills. 

A suitable solution for solving this problem is Hierarchical Task Networks (HTNs). As introduced in Section \ref{sec:intro}, HTNs are a formal planning framework for addressing complex problems by breaking them down into smaller hierarchically structured tasks.
HTN planners distinguish between two fundamental types of tasks: primitive and compound. Primitive tasks are the elemental building blocks of a plan, each representing a single executable step. In addition, primitive tasks are characterized by the operator, i.e., a list of agents capable of performing the task, a set of preconditions that must be met, and the effects of their execution. As such, the six actions introduced in our formalization can be implemented as primitive tasks within an HTN framework. In contrast, compound tasks result from the concatenation of several primitives. 

Moreover, HTNs' hierarchical structure complies with P5 as it provides explainable plans, which result in significant advantage when interacting with humans. Finally, this type of scheduler, together with the choice of state variables and the set of action templates illustrated above, makes it easy to achieve a modular architecture (P1) and runtime flexibility (P2). Regarding the modularity property, changes in the perception pipeline only require redefining a single action implementation. Moreover, the arrays of state features can accommodate varying numbers of agents, and it is possible to assign a restricted subset of actions to each agent by adjusting the preconditions of the actions. On the other hand, runtime flexibility is ensured by the planner's recursive nature, enabling dynamic adaptation to changing conditions.

\begin{figure*}[!ht]
\centering
\includegraphics[width=0.9\textwidth]{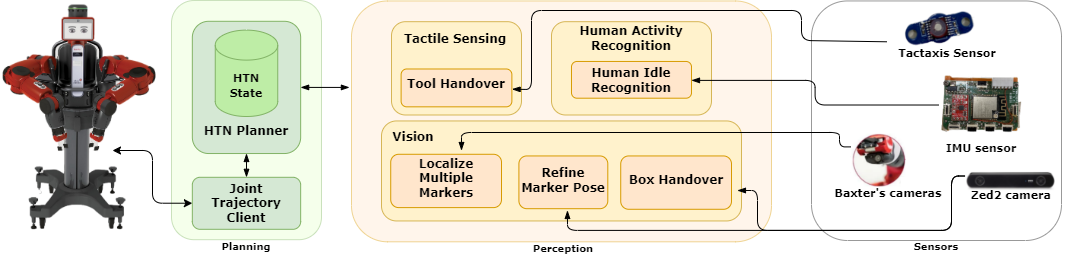}
\caption{Architecture diagram of the system. The HTN planner can activate the perception modules to update its state and move the robot using the Joint Trajectory Client. The perception module is composed of cameras and wearable and tactile sensors. The vision has three different modules: Localize Multiple Markers to identify and estimate the positions of the markers in the scene, Refine Marker Pose to improve the estimated position of a single marker before grasping, and Box Handover Detection to detect handover of small components such as screws. The wearables have been used to detect when the human is idle and the Tactile sensing to automatize the handover of tools using the shear forces.}
\label{fig:architecture}
\end{figure*}

\section{Implementation}
\label{Implementation}

In a real HRC scenario, the formalism introduced in the previous section can be used by a robot to describe the environment and determine which action to take accordingly. To this extent, alongside the HTN planning module, it is necessary to develop a perception system capable of recognizing the features that characterize the collaboration state.

In our implementation, the perception includes an external camera, two wrist-mounted robot cameras, a single tactile sensor on the robot gripper, and a set of wearable inertial measurement units (IMUs) monitoring human motion. We developed the appropriate software module to perceive a subset of the previously described state features. In particular, the external camera, together with the wrist-mounted cameras, is used to determine the positions of objects in the environment; this perception is simplified by using ArUco markers attached to the objects. In addition, wrist-mounted cameras have been adopted to determine the object characteristic (OC) (i.e., being empty or not) of some boxes containing small components. This way, through a simple colour segmentation module, the robot can determine whether the handled box is full or empty. Finally, the force sensor mounted on the gripper allows the robot to determine another property of a grasped object, i.e., if grasped by another agent. Regarding the agents' properties, the end-effector availability and pose of the robot are measured using its internal sensors, while the perception of the human relies on the IMU sensors. Specifically, users wear 4 inertial sensors, placed on the backs of their hands and under the wrist joint; the accelerations and angular velocities provided by the sensors are fed through an LSTM module to classify whether the person is idle or not (F1 score 98\%).
Therefore, the system has only a partial observation of the human agent pose and no information about their end-effector availability.

Given the described sensorial setup, the robot does not have complete knowledge of the state features. Furthermore, the Baxter robot is equipped with two simple grippers unsuitable for complex manipulation. Therefore, it can only perform actions belonging to the classes grasp, release, move, perceive, and wait. On the other hand, the human agent can perform all six activities, but most of them are not directly observed by the robot perception system. However, as stated in Section \ref{Experimental Setup}, the robot can empirically deduce which actions the human performed by observing the evolution of the environment, e.g., the motion of an object implies that an agent performed a specific set of actions.

As for the implementation of the actions, we distinguish the templates into two groups; the first one includes Grasp, Release, and Move actions, which directly influence the state variables of our system. The second comprises Perceive, which updates the planner's internal representation of the state, and Wait, which has been used to synchronize the collaboration. Considering our setup with a single stationary robot and two simple grippers, we designed only one implementation for each template in the first group. The Grasp changes the gripper width to exert a force onto an object. Its preconditions require the end-effector location to match the object location and the gripper end-effector availability (EEA) state variable to be true. This action sets EEA to false and updates the corresponding gripper joint state in the agent pose (AP) state variable. Release is the opposite operation as it opens the gripper releasing the object. The precondition is that EEA must be false, and the effects include setting EEA to true and updating the gripper joint state in AP just like in the previous case. Move describes the kinematic motion that leads the end-effector from location A to location B with a given orientation. The only precondition is that location B must be in the robot's reachable space, while the effect is the update of the agent pose (AP) and, in case the robot is holding an object, the object pose (OP). Similarly, Wait has a single implementation that pauses the planner until certain conditions are met. For example, in our implementation, Wait is used in conjunction with the tactile perception module to stop the execution until the human grasps the object held by the robot. As for the Perceive template, we provide five different implementations that refer to different sensors that constitute our perception pipeline. Additionally, these actions do not impact the state variables as they just update the internal representation of the HTN; therefore, they do not have any preconditions or effects. Check Available Objects, Precise Marker Detection and Detect Empty Box are the implementation related to the vision. The first one activates the ArUco detection and pose estimation from the stream of images obtained from the external camera and fills a list of available objects and their location. The second one refines the pose obtained by the previous action applying the same module on Baxter’s wrist camera and it is called when the robot arrives on top of a marker during a pick task. Lastly, the third one is activated when the robot performs the handover of a box and it employs color segmentation to track the box content. Detect Tool Pulling refers to the tactile perception module, it is called when the robot performs the handover of an object to monitor the behaviour of the shear forces sending a signal when an adaptive threshold is exceeded. Finally, Detect Idle activates the human activity recognition on the IMU sensors data stream firing a signal during the idle time.

\section{Experimental Setup}
\label{Experimental Setup}
The system described in the previous section was developed in Python, using the GTPyhop\cite{nau2021gtpyhop} library to implement the HTN planner. The workspace was framed using a Zed2 camera, and the collaborative robot Baxter by ReThink Robotics was utilized. For the perception of human activities, 4 MPU9250 IMU sensors\cite{carfì2024modular} were worn on the human hands and wrists, and a Tactaxis\cite{theo} sensor by Melexis\footnote{\scriptsize{\href{https://www.melexis.com/en}{melexis.com/en}}} was used for tactile perception. The robot's trajectories were planned using MoveIt!, the position and orientation of the ArUco markers were estimated using OpenCV, and the communication between modules was managed using ROS Noetic. A graphic representation of the architecture is provided in Fig.\ref{fig:architecture}, while the full code that drives these experiments is publicly available on our GitHub repository\footnote{\scriptsize{\href{https://github.com/TheEngineRoom-UniGe/HTN_Planner}{github.com/TheEngineRoom-UniGe/HTN\_Planner}}}.

Additionally, we developed a collaborative assembly scenario showcasing all proposed functionalities and the planner's ability to adapt to multiple consecutive interactions.
This scenario involves two people alternating in collaborating with Baxter to assemble one of four pieces of IKEA furniture. The available pieces of furniture are: i) an ODDVAR\footnote{\scriptsize{\href{https://www.ikea.com/it/it/p/oddvar-sgabello-pino-20249330/}{ikea.com/it/it/p/oddvar-sgabello-pino-20249330/}}} stool with 31 pieces, ii) a HUTTEN\footnote{\scriptsize{\href{https://www.ikea.com/it/it/p/hutten-portabottiglie-9-scomparti-legno-massiccio-70032451/}{ikea.com/it/it/p/hutten-portabottiglie-9-scomparti-legno-massiccio-70032451/}}} bottle rack with 24 pieces, iii) a KRITTER\footnote{\scriptsize{\href{https://www.ikea.com/it/it/p/kritter-seggiolina-bianco-40153699/}{ikea.com/it/it/p/kritter-seggiolina-bianco-40153699/}}} chair with 13 pieces and iv) a RÅGRUND\footnote{\scriptsize{\href{https://www.ikea.com/it/it/p/ragrund-porta-carta-igienica-bambu-30253072/}{ikea.com/it/it/p/ragrund-porta-carta-igienica-bambu-30253072/}}} paper roll stand with 16 pieces. To initiate interaction with the robot, each participant selects a piece of furniture, wears IMU sensors, places markers on assembly components, and positions them on tables near the robot. Once the preparation phase is complete, the planner waits for the user to be idle, adds the required actions to the plan to construct the chosen object, and begins execution. The human is aware of the action execution order and can refer to the instruction manual for the object, if necessary. The HTN plan for each piece of furniture is manually built using the primitives introduced in the previous section.

During runtime, the robot collects and hands the necessary components to the human. However, depending on the type of component, the transport modality differs. Using the perception pipeline, if larger components such as stool legs are detected, the robot retrieves them and places them on a table near the user. The user can then grab the objects, remove the markers, and proceed to assemble them. The human can also choose to pick up some of these items independently, and the perception modules will update the plan accordingly. The screwdriver and screw boxes are placed in predefined locations, where the robot will collect them and hold them in its gripper until the user needs them. The tactile sensor detects the handover of the screwdriver using shear forces, while a computer vision algorithm monitors container content through the robot's wrist-mounted camera. The user's primary objective is to assemble the components and secure the screws, whose number depends on the selected object. 
Some operations, such as connecting the legs of the stool to the seat and tightening all the screws, may require different times depending on the ability of the user. In that case, the framework relies on the human idle recognition to detect the end of the action. A video showing the whole assembly process is available on our YouTube channel\footnote{\scriptsize{\href{https://www.youtube.com/watch?v=Og_pvd0zKiU}{youtube.com/watch?v=Og\_pvd0zKiU}}}.

\section{Results}
\label{Results}
The experimental scenario showed the ability of the architecture to adapt to the assembly of multiple objects with variable numbers and types of components. 
Each trial comprehended the assembly of four pieces of furniture and lasted approximately 40 minutes. We repeated the process for four trials and the planner was never stopped for the entire duration of each trial. The only pauses were due to the waiting for the input with the name of the following object to be assembled. After subtracting the intervals required for component alignment and sensor exchange between the experimenters, who are the only two users involved, the net average duration of the collaborative activity is 29 minutes and 28 seconds. Additionally, the average time needed to build each object was 7 minutes and 31 seconds (SD=1.84). The assembly of the stool was the longest with 8 minutes and 25 seconds, while all the other objects had an average assembly time close to 6 minutes and a half. 

\begin{figure}[!t]
\centering
\includegraphics[width=0.45\textwidth]{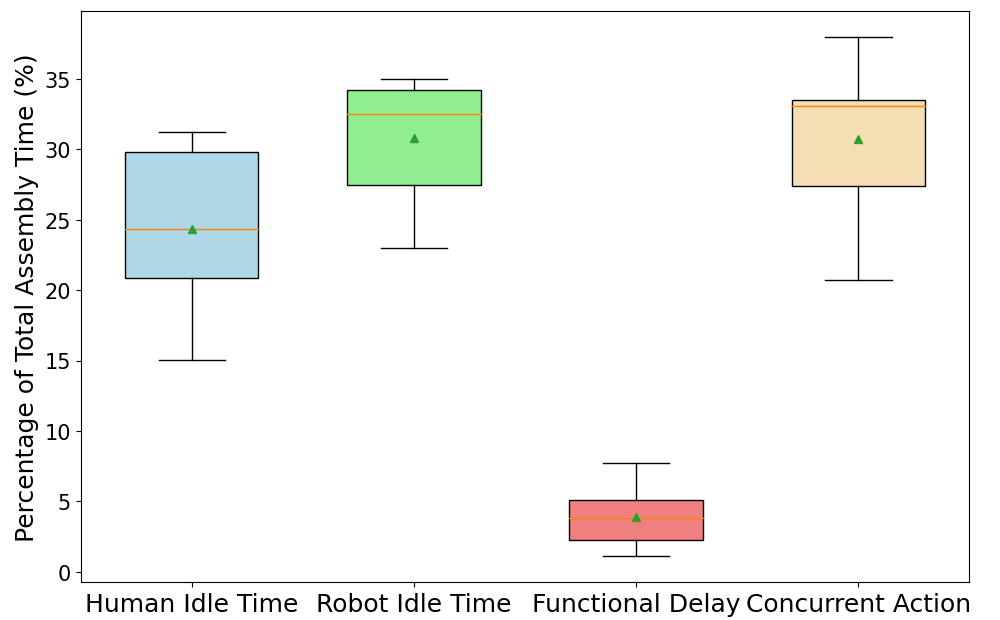}
\caption{The plots show the fluency metrics expressed as a percentage of the assembly time. From left to right we provide human idle time, robot idle time, functional delay and concurrent action time.}
\label{fig:idletime}
\end{figure}

\begin{figure*}[!t]
\centering
\includegraphics[width=0.9\textwidth]{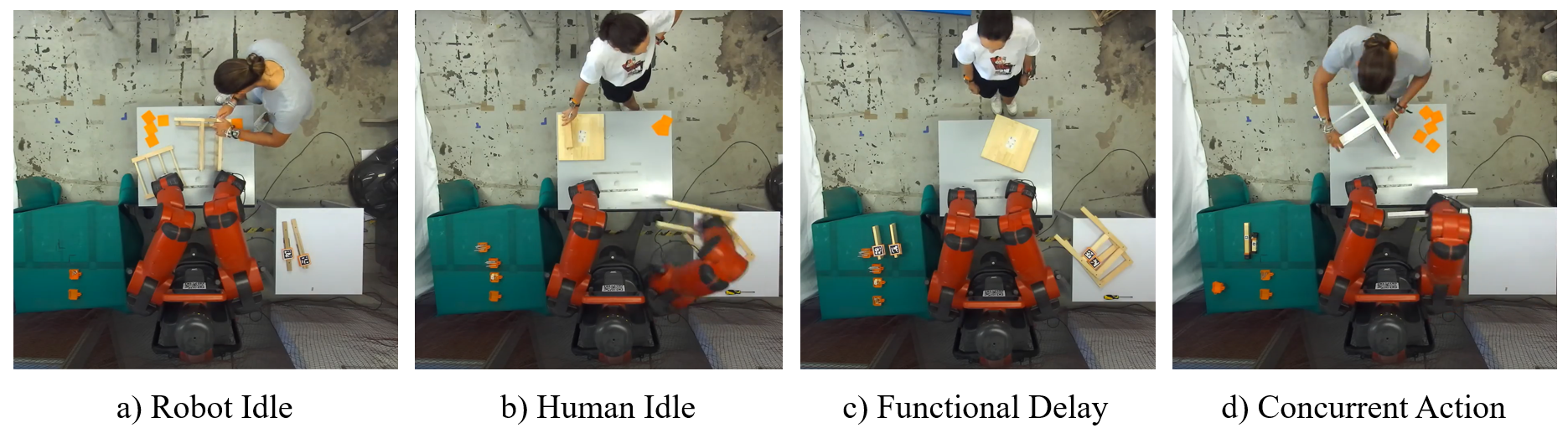}
\caption{The picture represents four frames from the experimental scenario referring to the four fluency metrics. The frames respectively represent Robot Idle time (a), Human Idle time (b), Functional Delay (c) and Concurrent Action (d).}
\label{fig:frames}
\end{figure*}

This result is influenced by the number of actions included in the planning, which is 545 for the stool, 385 for the bottle holder, 322 for the small chair and 195 for the paper stand. Similarly, the planning time is directly proportional to the number of actions; therefore, the stool takes the longest at 0.23 seconds and the paper stand is the quickest at 0.08 seconds. These delays are negligible compared to the time required to complete the task, therefore, we can conclude that the scheduler does not have a significant effect on the performance of the framework, adhering to P4. 
Fig. \ref{fig:idletime} shows the fluency metrics evaluated in our scenario. The results show that the robot's idle time and concurrent action occupy most of the total assembly time. Specifically, robot idle time has the highest average with 30.83\% and concurrent action follows with 30.74\%. The second category, however, has a larger spread and manages to reach almost 40\% in some instances. As for the human's idle time, it has a lower average of 24.33\% and the overlap time in which both agents are idle is very low with 3.86\% on average. Fig. \ref{fig:frames} shows an example for each of the four categories from the experimental scenario.

\begin{figure}[!t]
\centering
\includegraphics[width=0.5\textwidth]{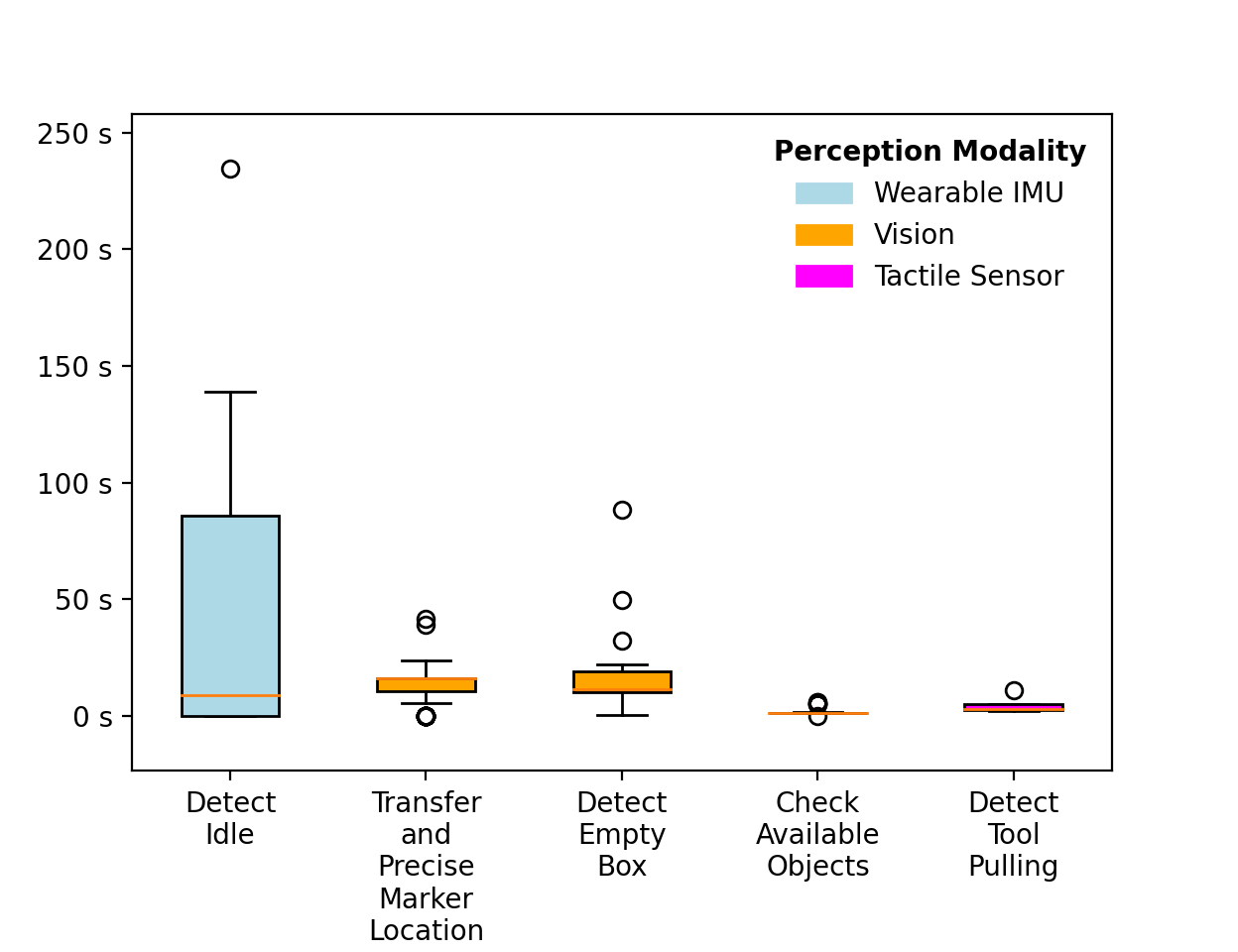}
\caption{Time needed for each action involving perception during the collaborative scenario. The different colours refer to the perception modality associated with each action.}
\label{fig:perceptiontime}
\end{figure}

It is necessary to observe that these percentages are not only influenced by the framework but also by the structure of the task. In addition, robot idle time includes all actions related to perception. Therefore, a large part of the robot's idle time is due to awaiting human actions in assembly. Additionally, In the current system implementation, the robot is limited to performing only one action at a time and using both robot arms in parallel for collecting the objects could significantly decrease the human waiting. Finally, regarding the spread of the data shown in the boxplots, we should point out that the framework allows adjustment of the plan at runtime to accommodate user preferences and re-plan in case of an error. This feature, which introduces variability in action coordination, corresponds to point P2 in the list of properties defined in the introduction, and we show an example of it in a second video on our YouTube channel\footnote{\scriptsize{\href{https://www.youtube.com/watch?v=c2YShK02fsI&ab_channel=TheEngineRoom}{youtube.com/watch?v=c2YShK02fsI}}}.

Referring to the function implementations provided in Section \ref{Implementation}, we extrapolated the time required to execute each of them, and the results are represented in Fig.\ref{fig:perceptiontime}. The data shown only pertain to perception-related actions, as these are the only ones that vary according to the behaviour of the two agents and the state of the plan. It can be seen that the ``Detect Idle'', ``Detect Tool Pulling'', and ``Detect Empty Box'' activities, which need active user participation, take longer and have a higher variation. ``Detect Idle'', in particular, has a significantly greater duration since it is typically employed while waiting for the human to finish assembly operations. The sole exception is ``Transfer and Precise Marker Location'', which has a variable duration although it does not require interaction between the two agents. However, in our implementation, this action includes both the movement of the robotic arm and the refinement of the marker pose; therefore, most of the variability for this action can be explained by the fact that the time needed to complete each trajectory depends on its length. Finally, ``Wait'' was not reported in the graph because, in our implementation, it always takes place in conjunction with a perception-related action, and its length correlates with the data already shown.

\section{Conclusions}
In conclusion, our study has introduced a flexible framework for Human-Robot Collaboration able to adapt to different scenarios, while adhering to five key properties: Modular Architecture (P1), Runtime Flexibility (P2), Extensible Task Description (P3), Low Computational Overhead (P4) and Interpretability (P5). Interpretability (P5) is inherently derived from the hierarchical nature of our HTN planner, while P1 and P3 are additionally shaped by the selection of specific action templates, as outlined in section \ref{Formalism}. To address P2, we designed action implementations that facilitate error recovery and adjust to diverse human behaviours, further illustrated through a video demonstration. Additionally, we provided metrics for the planning time that evidence P4 is preserved in our framework.

In the experimental scenario, two human agents alternated in collaborating with a robot to assemble objects of varying shapes and complexity. Notably, the collaborative process proceeded seamlessly without interruptions in the planning process achieving satisfactory fluency metrics. However, our findings indicated that there is room for improvement in the system. Firstly, in terms of perception capabilities, we recognize the importance of avoiding scripted locations for objects such as screwdrivers and boxes. Additionally, enhancing human action recognition beyond simple idle and active states and using the results of this recognition as preconditions for other actions could further refine our system. Finally, at present, the construction of the plan is manually done offline and automating this process could greatly improve the versatility of the framework. 

Moreover, we observed significant idle time for both human and robot agents, with the duration of perception actions primarily dependent on human decisions. As a promising avenue for future work, we suggest exploring the potential for parallel execution by involving multiple human agents simultaneously, especially when tasks such as assembling chairs can benefit from utilizing both arms of the robot concurrently. This approach promises to achieve even greater efficiency in future HRI applications, as we continue to refine and expand upon our formalization framework and plan to test it with a broader sample of participants. Additionally, while our framework provides runtime flexibility, it currently does not support dynamic task sequence adjustments, which are critical for effectively handling significant deviations by human agents during collaborative activities. Future research should focus on enabling these higher-level task adjustments to improve collaboration efficiency and adaptability.

\section*{Acknowledgement}
The authors would like to acknowledge Melexis\footnotemark[2] for providing the Tactaxis sensor, which constitutes the tactile perception of the proposed architecture. Moreover, this research was partially supported by the Italian government under the National Recovery and Resilience Plan (NRRP), Mission 4, Component 2 Investment 1.5, funded from the European Union NextGenerationEU and awarded by the Italian Ministry of University and Research.

\bibliographystyle{IEEEtran}

\bibliography{bibliography}

\end{document}